\documentclass[11pt,a4paper]{article}
\usepackage[hyperref]{eacl2021}
\usepackage{fancyhdr,graphicx,amsmath,amssymb}
\usepackage{times}
\usepackage{latexsym}

\usepackage{multirow}
\usepackage{microtype}

\aclfinalcopy % Uncomment this line for the final submission
\title{Sentiment Classification in Swahili Language Using Multilingual BERT}

\author{Gati L. Martin \\
  Soonchunhyang University, \\
  Asan, South Korea \\
  \texttt{gatimartin@sch.ac.kr} \\\And
  Medard E. Mswahili \\
  Soonchunhyang University, \\
  Asan, South Korea \\
  \texttt{medardedmund25@sch.ac.kr} \\\AND
  Young-Seob Jeong \\
  Soonchunhyang University, \\
  Asan, South Korea \\
  \texttt{bytecell@sch.ac.kr} \\
  }
 
\date{}

\begin{document}
\maketitle

\begin{abstract}
The evolution of the Internet has increased the amount of information that is expressed by people on different platforms.
This information can be product reviews, discussions on forums, or social media platforms.
Accessibility of these opinions and people’s feelings open the door to opinion mining and sentiment analysis. 
As language and speech technologies become more advanced, many languages have been used and the best models have been obtained. 
However, due to linguistic diversity and lack of datasets, African languages have been left behind.
In this study, by using the current state of the art model, multilingual BERT, we perform sentiment classification on Swahili datasets.
The data was created by extracting and annotating 8.2k reviews and comments on different social media platforms and ISEAR emotion dataset.
The data were classified as either positive or negative. 
The model was fine-tuned and achieve the best accuracy of 87.59\%.  
\end{abstract}

\section{Introduction}
The growth of the Internet has led to the increase in data that holds infinite and valuable perceptions about public opinion. 
Since the amount of generated data is too large for the normal users to analyze, sentiment analysis techniques are used to automate the process.
Sentiment classification deals with identifying and classifying opinions in the text using natural language processing (NLP) techniques.
Sentiment analysis has been popular in different applications like customer feedback analysis, social media monitoring, and product and services analysis.
The result obtained can be useful for understanding user's perceptions and satisfaction toward products and services.

Sentiment classification has evolved with different machine learning techniques including traditional machine learning~\citep{samuel2020covid} and deep learning~\citep{kim2019sentiment} techniques.
Recently, significant results have been reported by using pretrained state-of-the-art, BERT (Bidirectional Encoder Representations from Transformers) model~\citep{devlin2018bert}. 
Many researchers have implemented bert-based model on different NLP tasks, including sentiment classification, intent detection, and emotion classification.
The majority of these studies have either focused on English language or high resource languages.
However, we lack these implementations on low resource languages such as Swahili, Yoruba, and Zulu.

Swahili is a Bantu language spoken in multiple countries in Africa but mainly in Tanzania, Kenya, and Uganda as the official language. 
It contains many loanwords from Arabic, English, and other Bantu languages. 
Africa is approximately one-third of the world's languages, it has a language diversity of over 2,000 languages, many of which are primarily oral, and little is written.
This shortage of online resources and datasets keeps the researches in these geographical areas to be stunted despite the free availability of NLP architectures.

In this study we perform binary sentiment classification, using multilingual BERT (mBERT) by creating an 8.2k Swahili dataset extracted from different Swahili online platforms such as JamiiForum,\footnote{https://www.jamiiforums.com/} and DW Kiswahili (Deutsche Welle).\footnote{https://www.facebook.com/dw.kiswahili}

\section{Related Work}
Sentiment classification is one of the most popular tasks in NLP, therefore, there has been a lot of researches conducted using different machine learning techniques.
These researches have been focused on either binary (positive and negative) or ternary (positive, negative, and neutral) sentiment classification.
Study of ~\citet{jagdale2019sentiment} use support vector machine (SVM) and Naïve Bayes (NB) techniques for binary classification of Amazon product reviews and achieve 98.17\% and 93.54\% accuracy for camera reviews.
~\citet{samuel2020covid} conducted a binary classification on COVID-19 Tweets to understand COVID-19’s informational crisis by comparing two essential ML methods in the context of textual analytic. 
NB achieve an accuracy of 91\% compared to the accuracy of 74\% of Logistic regression for short Tweets, and poor performance for both in longer Tweets. 
Although these traditional machine learning (ML) models have shown great contributions, they suffer from the limitation of relying on feature selection for their performance. 
Features are defined and extracted either manually or by using feature selection methods.

The deep learning (DL) technique is known for its competence in extracting the features automatically.
~\citet{kumar2020exploring} compared ML (maximum entropy, NB, SVM) and DL approaches (long short term memory (LSTM)~\citep{hochreiter1997long}, convolution neural network (CNN))~\citep{kim2019convolutional} for exploring the impact of age and gender on the binary classification of sentiment reviews, and CNN achieves the best accuracy of 78\% on age and 80\% on gender impacts.
~\citet{li2020textual} adopted simple recurrent network (SRN), LSTM, and CNN to sentiment analysis tasks on movie reviews dataset for evaluating the effect of data quality on model performance. In this study, CNN achieve higher accuracy with short and readable reviews. 
~\citet{kim2019sentiment} use three movie review data to designed a binary and ternary sentiment classification model of accuracy 81\% and 68\% and shows that employing consecutive convolutional layers is effective for longer texts.
~\citet{dang2020sentiment} examine two text processing techniques, word embedding and term frequency-inverse document frequency (TF-IDF) on deep neural network, recurrent neural network, and CNN models by using 8 datasets (tweets and reviews), results show that it is better to combine deep learning techniques with word embedding than with TF-IDF. 

Some pre-trained word embedding such as Word2Vec~\citep{mikolov2013efficient} face some limitation of incapability to handle out-of-vocabulary and being context-independent.     
A pretrained state-of-the-art model, BERT~\citep{devlin2018bert} handle these limitations by using attention mechanism that takes the context into consideration. 
This model has achieved incredible results in many NLP tasks including sentiment classification.
~\citet{lee2020bert} perform binary sentiment classification on U.S. stocks reviews and achieve an accuracy of 87.3\%.  
The study of~\citet{9151169} use the Chinese BERT model to classify ternary sentiments and analyze the characteristics of negative sentiment about COVID-19 in a popular Chinese social media (Weibo).
~\citet{biswas2020achieving} analyze 3 sentiments of sentences in Stack Overflow posts using BERT and achieve an F1 score of 87\%.

There are studies that show the great impact of mBERT on low resource languages.
~\citet{messaoudi2020learning} compared binary classification of several deep learning models on 9k Tunisian social media comments and achieved the accuracy of 93.8\% using mBERT.
~\citet{cruz2019evaluating} evaluate fine-tuning techniques (BERT and Universal Language Model Fine-tuning~\citep{howard2018universal}) on binary sentiment data of Filipino language, and achieve better performance using BERT.
According to our knowledge, there is no BERT-based model for sentiment classification that has implemented in the Swahili language.
By using mBERT we perform the binary sentiment classification on Swahili annotated data.

\section{Method}
\begin{figure}[ht]
	\centering
		\includegraphics[scale=0.7]{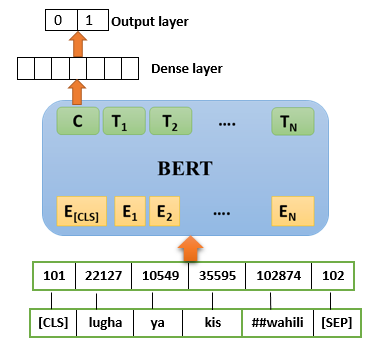}
	\caption{Model architecture}
	\label{ModelA}
\end{figure}

BERT is a neural network-based language model developed by Google ~\citep{devlin2018bert}. BERT is bidirectional, unsupervised language representation, pre-trained using only a plain BooksCorpus (800M words) and English Wikipedia (2,500M words).
This characteristic allows the model to learn the context of a word based on all of its surrounding text.
There are two architecture sizes, BERT Base and BERT Large with 12 and 24 encoder layers respectively.
The model has two phases: pre-training and fine-tuning. 
During pre-training, the model is trained on unsupervised data over different pre-training tasks. 
For fine-tuning, the model is first initialized with the pre-trained parameters and then fine-tuned using labeled data from the downstream tasks.
The model takes input data in a specific format with limited input sequences of up to 512 tokens. 
Input to BERT can be a single sentence or a sentence pair with special tokens [CLS] to indicate the beginning of the sentence and [SEP] to indicate the end of the separation of sentences.
To feed our sentences to BERT, they must be split into tokens by using the WordPiece tokenizer, and then these tokens must be mapped to their index in the tokenizer vocabulary.
BERT uses the WordPiece algorithm to generate a fixed-size vocabulary of individual characters, most common words, and subwords in a trained language corpus.
In fixed vocabulary some tokens might not appear and cause out-of-vocabulary(OOV) issue, this is where WordPiece algorithm takes effect by split them into character tokens that can be mapped in the vocabulary file and assign \#\# to indicate that it is a suffix following some other subwords (as shown in Figure \ref{ModelA}).

In our experiments, we use a multilingual version of BERT (mBERT) that is trained on the Wikipedia pages of 104 languages, with a shared word piece vocabulary. 
Swahili is among the language that mBERT was trained in, this adds some advantages of using the model to encounter the linguistic features of the language. 
Tanzania has many tribes (Sukuma, Konde, Maasai) that differ by their accents, this has affected the representation of the Swahili language in both oral and written form.
While most of conducted researches are based on standard (formal) language, the most used language in the social media platforms is informal language (Table \ref{table informal}). 
This has been influenced by age difference ~\citep{kumar2020exploring} and loanwords from other languages.
There are English words that are modified and used as Swahili but do not contain semantic meaning, for example, ccta (sister) means dada, or faza (father) means baba.
All of these features were involved in our dataset because the social media platforms involve people of different age, location and aspects.
These characteristics, make mBERT the better choice to handle the OOV words and accommodate the loanwords from other languages.
Fig. \ref{ModelA} shows the BERT model architecture that we use in fine-tuning the sentiment classification.
We add a dense layer on top of the pre-trained model and maintain other hyper-parameters as stated in the original paper.

\begin{table}[ht!]
\centering
\scalebox{0.8}{%
\begin{tabular}{c c c}
\hline & \textbf{Positive} & \textbf{Negative}\\
\hline
informal & manzi mkali & miyeyusho kinoma\\
formal & binti mzuri  & tenda kinyume \\
English & georgeous lady & disappointment \\
\hline
\end{tabular}}
\caption{Informal Sentiment example}
\label{table informal}
\end{table} 

\section{Experiments}
\subsection{Dataset}
We use ISEAR~\footnote{\url{https://www.kaggle.com/faisalsanto007/isear-dataset}} emotional dataset which contains 7 emotions (e.g., joy, anger, sadness, disgust, shame, and guilt), and convert them into sentiments by taking joy as positive sentiment and all others as negative.
We use other data from different sources including online discussion forums (e.g., JamiiForum, DW Kiswahili) and social media (e.g., Tweeter, YouTube), and manually annotated them; we finally have a total of 8.2k data including 2.7k positive and 5.4k negative sentiments.
The extraction of data was done manually except for the tweets where we use Tweepy, the open source Python library to access tweeter API. \footnote{\url{https://developer.twitter.com/en/docs/twitter-api}}
The dataset cover different topics including politics, different aspects of daily life activities and psychology (ISEAR).

Annotation process was done by two native Swahili speakers from Tanzania. 
Each annotator has to identify the sentiment of the text then, labelled data were compared. 
For those with different labels, the annotators have to agree on the correct label without uncertainty.
We remove all the text which the labelling was not certain and the neutral ones.
Social media data are very noisy, they include many unstructured data that may affect the model performance~\cite{li2020textual}. 
To be on the safe side, we pre-process our data to remove unwanted characters and words such as usernames, links, and some punctuation.
We split the data into two sets, where 10\% was used for testing and the remaining 90\% for the training process.

\begin {table}[ht]
\centering
\begin{tabular}{c c} 
\hline
 Number of text & 8,218 \\
 Annotated data & 4,938 \\
 Number of annotator & 2  \\
 ISEAR & 40\% \\
 JamiiForums & 29\% \\
 DW Kiswahili & 2\% \\
 Tweeter & 18\% \\
 YouTube & 10\% \\
\hline
\end{tabular}
\caption{Dataset statistics}
\label{dataset_table}
\end{table}

\subsection{Hyper-parameters}
To fine-tune for the sentiment classification task, we use the HuggingFace PyTorch library\footnote{\url{https://huggingface.co/transformers/model_doc/bert.html}} to initialize the model with the pre-trained parameters.
Pre-training was done on GeForce GTX 1080 Ti GPU, with a batch size of 16 and a maximum sequence length of 64. 
Other parameters remain as defined in BERT pre-training, Adam optimizer~\citep{kingma2014adam} with a learning rate of 5e-5 and an epsilon parameter of 1e-8. 

\begin {table}[h!]
\centering
\begin{tabular}{c c c c} 
\hline
 \textbf{Sentiment} &\textbf{Precision} & \textbf{Recall} & \textbf {F1 score} \\
\hline
 Negative & 0.88 & 0.94 & 0.91\\
 Positive & 0.87 & 0.76 & 0.81  \\
\hline
\end{tabular}
\caption{Result based on sentiments}
\label{Result_table}
\end{table}

\begin {table}[ht]
\centering
\begin{tabular}{c c c c } 
\hline
 & & \multicolumn{2}{c}{Predicted} \\
\hline
  & & Negative & Positive \\
\multirow{2}{3em}{Actual} & Negative & 501 & 34 \\
                          & Positive & 68 & 219 \\
\hline
\end{tabular}
\caption{Confusion matrix table}
\label{confusion_matrix}
\end{table}

\subsection{Result}
Table \ref{Result_table} shows the sentimental classification result of our dataset.
The accuracy metrics, precision, recall, and F1-score of each sentiment ware computed.
Our model achieves an accuracy of 87.59\%.
Based on the result we can see that the scores of negative sentiment are high compared to positive sentiments.
Table \ref{confusion_matrix} shows the confusion matrix, where the ratio of positive text that were predicted to be negative was high. 
This was due to an unbalanced ratio of data, a high ratio of negative data gives the model high sensitivity on negative predictions.

\section{Conclusion}
In this paper, we perform the sentiment classification on the Swahili dataset that we extract from different online social media platforms. 
We apply the pre-trained mBERT model and achieve the best accuracy of 87.59\%.
As stated above, there are some biases in free available social media platforms data that can contribute to sarcasm, poor data quality, and semantic problems that can affect the model performance.
In addition to observations that we found on predictions, the size and ratio of the data is important for model performance. However,this research is in-progress the following issues will be addressed, the ratio of data and comparison with other NLP models.
While mBERT has shown great performance, there are other studies that demonstrate the powerful ability of language-specific BERT models over mBERT.
In the future, we will train specific BERT for Swahili that can be favorable to language features and applicable to different tasks.

%\bibliography{anthology,eacl2021}
\bibliography{eacl2021}

\begin{thebibliography}{17}
\expandafter\ifx\csname natexlab\endcsname\relax\def\natexlab#1{#1}\fi

\bibitem[{Biswas et~al.(2020)Biswas, Karabulut, Pollock, and
  Vijay-Shanker}]{biswas2020achieving}
Eeshita Biswas, Mehmet~Efruz Karabulut, Lori Pollock, and K~Vijay-Shanker.
  2020.
\newblock Achieving reliable sentiment analysis in the software engineering
  domain using bert.
\newblock In \emph{2020 IEEE International Conference on Software Maintenance
  and Evolution (ICSME)}, pages 162--173. IEEE.

\bibitem[{Cruz and Cheng(2019)}]{cruz2019evaluating}
Jan Christian~Blaise Cruz and Charibeth Cheng. 2019.
\newblock Evaluating language model finetuning techniques for low-resource
  languages.
\newblock \emph{arXiv preprint arXiv:1907.00409}.

\bibitem[{Dang et~al.(2020)Dang, Moreno-Garc{\'\i}a, and De~la
  Prieta}]{dang2020sentiment}
Nhan~Cach Dang, Mar{\'\i}a~N Moreno-Garc{\'\i}a, and Fernando De~la Prieta.
  2020.
\newblock Sentiment analysis based on deep learning: A comparative study.
\newblock \emph{Electronics}, 9(3):483.

\bibitem[{Devlin et~al.(2018)Devlin, Chang, Lee, and
  Toutanova}]{devlin2018bert}
Jacob Devlin, Ming-Wei Chang, Kenton Lee, and Kristina Toutanova. 2018.
\newblock Bert: Pre-training of deep bidirectional transformers for language
  understanding.
\newblock \emph{arXiv preprint arXiv:1810.04805}.

\bibitem[{Hochreiter and Schmidhuber(1997)}]{hochreiter1997long}
Sepp Hochreiter and J{\"u}rgen Schmidhuber. 1997.
\newblock Long short-term memory.
\newblock \emph{Neural computation}, 9(8):1735--1780.

\bibitem[{Howard and Ruder(2018)}]{howard2018universal}
Jeremy Howard and Sebastian Ruder. 2018.
\newblock Universal language model fine-tuning for text classification.
\newblock \emph{arXiv preprint arXiv:1801.06146}.

\bibitem[{Jagdale et~al.(2019)Jagdale, Shirsat, and
  Deshmukh}]{jagdale2019sentiment}
Rajkumar~S Jagdale, Vishal~S Shirsat, and Sachin~N Deshmukh. 2019.
\newblock Sentiment analysis on product reviews using machine learning
  techniques.
\newblock In \emph{Cognitive Informatics and Soft Computing}, pages 639--647.
  Springer.

\bibitem[{Kim and Jeong(2019)}]{kim2019sentiment}
Hannah Kim and Young-Seob Jeong. 2019.
\newblock Sentiment classification using convolutional neural networks.
\newblock \emph{Applied Sciences}, 9(11):2347.

\bibitem[{Kim(2019)}]{kim2019convolutional}
Y~Kim. 2019.
\newblock Convolutional neural networks for sentence classification. arxiv
  2014.
\newblock \emph{arXiv preprint arXiv:1408.5882}.

\bibitem[{Kingma and Ba(2014)}]{kingma2014adam}
Diederik~P Kingma and Jimmy Ba. 2014.
\newblock Adam: A method for stochastic optimization.
\newblock \emph{arXiv preprint arXiv:1412.6980}.

\bibitem[{Kumar et~al.(2020)Kumar, Gahalawat, Roy, Dogra, and
  Kim}]{kumar2020exploring}
Sudhanshu Kumar, Monika Gahalawat, Partha~Pratim Roy, Debi~Prosad Dogra, and
  Byung-Gyu Kim. 2020.
\newblock Exploring impact of age and gender on sentiment analysis using
  machine learning.
\newblock \emph{Electronics}, 9(2):374.

\bibitem[{Lee et~al.(2020)Lee, Gao, and Tsai}]{lee2020bert}
Chien-Cheng Lee, Zhongjian Gao, and Chun-Li Tsai. 2020.
\newblock Bert-based stock market sentiment analysis.
\newblock In \emph{2020 IEEE International Conference on Consumer
  Electronics-Taiwan (ICCE-Taiwan)}, pages 1--2. IEEE.

\bibitem[{Li et~al.(2020)Li, Goh, and Jin}]{li2020textual}
Lin Li, Tiong-Thye Goh, and Dawei Jin. 2020.
\newblock How textual quality of online reviews affect classification
  performance: a case of deep learning sentiment analysis.
\newblock \emph{Neural Computing and Applications}, 32(9):4387--4415.

\bibitem[{Messaoudi et~al.(2020)Messaoudi, Haddad, HajHmida, Fourati, and
  Hamida}]{messaoudi2020learning}
Abir Messaoudi, Hatem Haddad, Moez~Ben HajHmida, Chayma Fourati, and
  Abderrazak~Ben Hamida. 2020.
\newblock Learning word representations for tunisian sentiment analysis.
\newblock \emph{arXiv preprint arXiv:2010.06857}.

\bibitem[{Mikolov et~al.(2013)Mikolov, Chen, Corrado, and
  Dean}]{mikolov2013efficient}
Tomas Mikolov, Kai Chen, Greg Corrado, and Jeffrey Dean. 2013.
\newblock Efficient estimation of word representations in vector space.
\newblock \emph{arXiv preprint arXiv:1301.3781}.

\bibitem[{Samuel et~al.(2020)Samuel, Ali, Rahman, Esawi, Samuel
  et~al.}]{samuel2020covid}
Jim Samuel, GG~Ali, Md~Rahman, Ek~Esawi, Yana Samuel, et~al. 2020.
\newblock Covid-19 public sentiment insights and machine learning for tweets
  classification.
\newblock \emph{Information}, 11(6):314.

\bibitem[{{Wang} et~al.(2020){Wang}, {Lu}, {Chow}, and {Zhu}}]{9151169}
T.~{Wang}, K.~{Lu}, K.~P. {Chow}, and Q.~{Zhu}. 2020.
\newblock \href {https://doi.org/10.1109/ACCESS.2020.3012595} {Covid-19
  sensing: Negative sentiment analysis on social media in china via bert
  model}.
\newblock \emph{IEEE Access}, 8:138162--138169.

\end{thebibliography}
\bibliographystyle{acl_natbib}

\end{document}